\documentclass[10pt,twocolumn,letterpaper]{article}

\usepackage{cvpr}              
\definecolor{cvprblue}{rgb}{0.21,0.49,0.74}
\usepackage[pagebackref,breaklinks,colorlinks,allcolors=cvprblue]{hyperref}
\usepackage{graphicx}
\usepackage{amsmath}
\usepackage{amssymb}
\usepackage{booktabs}
\usepackage{enumitem}
\usepackage{multirow}
\def\paperID{4254} 
\def\confName{CVPR}
\def\confYear{2026}

\title{Seeing What Matters: Visual Preference Policy Optimization for Visual Generation}

\author{
	Ziqi Ni$^{1}$\thanks{Equal contribution. Work done when Ziqi interned at Institute of Artificial Intelligence (TeleAI), China Telecom.},\ 
	Yuanzhi Liang$^{2*}$,\
	Rui Li$^{2,3}$,\
	Yi Zhou$^{1}$\thanks{Corresponding authors.},\
	Haibin Huang$^{2}$, \
	Chi Zhang$^{2}$,\
	Xuelong Li$^{2\dagger}$\\
	$^{1}$Southeast University, $^{2}$Institute of Artificial Intelligence (TeleAI), China Telecom\\
	$^{3}$University of Science and Technology of China\\
	{\tt\small zqni@seu.edu.cn, liangyzh18@outlook.com, yizhou.szcn@gmail.com, xuelong\_li@ieee.org}
}

\begin{document}
\maketitle
\begin{abstract}
Reinforcement learning (RL) has become a powerful tool for post-training visual generative models, with Group Relative Policy Optimization (GRPO) increasingly used to align generators with human preferences. However, existing GRPO pipelines rely on a single scalar reward per sample, treating each image or video as a holistic entity and ignoring the rich spatial and temporal structure of visual content. This coarse supervision hinders the correction of localized artifacts and the modeling of fine-grained perceptual cues. We introduce Visual Preference Policy Optimization (ViPO), a GRPO variant that lifts scalar feedback into structured, pixel-level advantages. ViPO employs a Perceptual Structuring Module that uses pretrained vision backbones to construct spatially and temporally aware advantage maps, redistributing optimization pressure toward perceptually important regions while preserving the stability of standard GRPO. Across both image and video benchmarks, ViPO consistently outperforms vanilla GRPO, improving in-domain alignment with human-preference rewards and enhancing generalization on out-of-domain evaluations. The method is architecture-agnostic, lightweight, and fully compatible with existing GRPO training pipelines, providing a more expressive and informative learning signal for visual generation.
\end{abstract}    
\section{Introduction}
\label{sec:intro}
Reinforcement learning (RL) has recently emerged as an effective framework for aligning visual generative models~\cite{diffusion_ho,diffusion_song,flow_matching,rectifiedflow,liang2026integrating,zhang2024vast,aiflow,teleboost} with human preferences~\cite{firstrlddpm,fan2023dpok}, enabling scalable supervision beyond paired data. Among RL-based approaches, Group Relative Policy Optimization (GRPO)~\cite{deepseek} has attracted attention for its group-wise comparison-based advantage formulation, which improves optimization stability and sample quality. Recent studies~\cite{dancegrpo,zhou2024flowgrpo} have successfully extended GRPO to diffusion and flow-based generators, confirming its potential for reinforcement-driven alignment in visual generation.

However, GRPO was originally designed for token-level or sequence-level outputs, such as in language or reasoning tasks. When directly applied to visual data, this formulation assumes that each visual instance, whether a static image or a video, can be represented by a single scalar advantage, ignoring the rich spatial and temporal structure inherent in visual generation. Such simplification makes GRPO less sensitive to regional or semantic variations within visual content, limiting its ability to assign differentiated credit across spatial locations. Consequently, although the framework remains effective in principle, it provides insufficiently structured feedback for complex visual synthesis tasks.
\begin{figure}
	\centering
	\setlength{\belowcaptionskip}{0pt}
	\includegraphics[width=\linewidth]{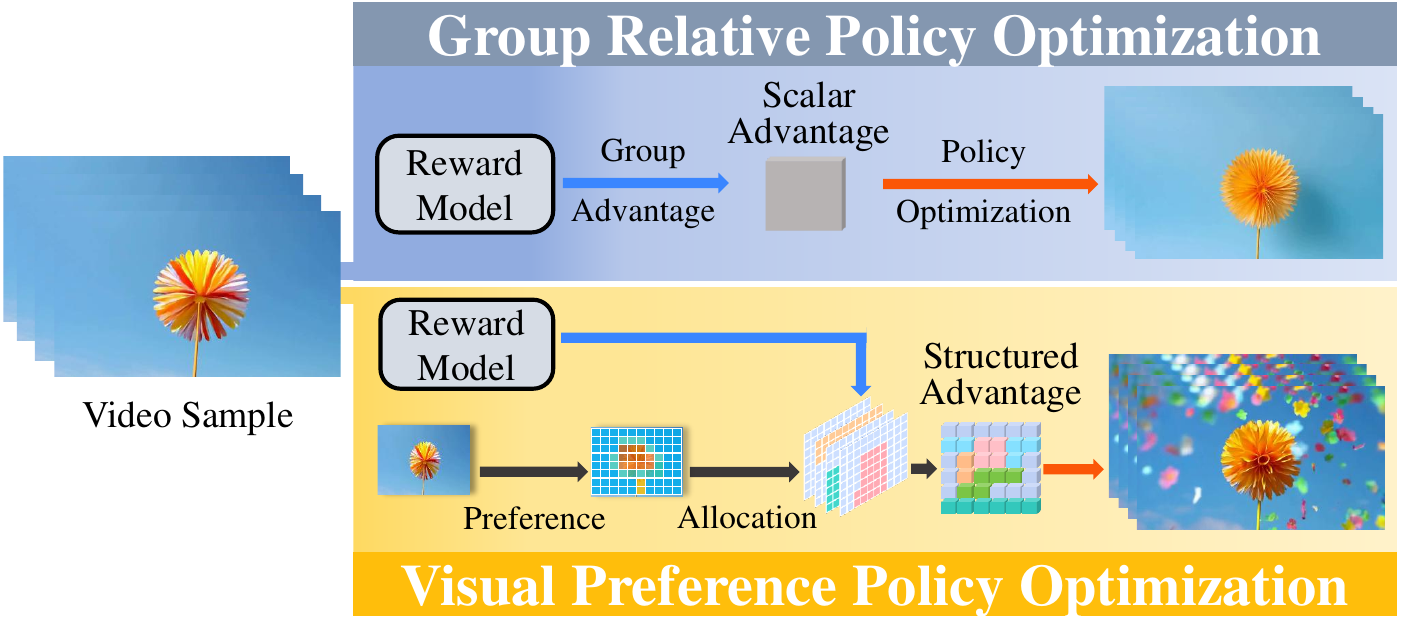}
	\caption{Brief illustration of our work. Existing GRPO for visual generation assigns a single scalar advantage to the entire content, producing coarse feedback that often leads to sub‑optimal results. In contrast, our ViPO converts this coarse signal into preference‑aware feedback, enabling fine‑grained alignment. This allows, for instance, differentiated optimization of the dancing doll and its background, yielding outputs that are more coherent, harmonious, and perceptually pleasing.}
	\label{fig:motive}
\end{figure}
Specifically, this coarse feedback directly affects the visual quality and perceptual alignment of generated results. In conventional GRPO, all pixels share an identical scalar advantage, implying uniform contribution to perceptual quality. 
This uniform weighting disregards the varying contributions of different regions to perceptual quality, producing indiscriminate gradients that can amplify irrelevant or misleading cues,
as illustrated in Figure~\ref{fig:motive}. 
This reflects a 
spatial credit assignment problem in RL, where undifferentiated rewards misguide optimization and limit the generator’s capacity to produce perceptually faithful and semantically consistent outputs. 
These limitations motivate the need for a fine-grained, perception-guided policy optimization framework specifically designed for visual content generation.

To overcome these limitations, we introduce Visual Preference Policy Optimization (ViPO), a redesigned GRPO framework for visual content generation. 
ViPO reformulates the advantage representation and introduces spatial credit allocation, enabling differentiated feedback across perceptually distinct regions.
adapting the original GRPO to better handle the structured feedback required in image and video generation. It transforms the coarse scalar advantage into structure-aware feedback guided by perceptual embeddings. Instead of applying a single scalar advantage to the whole sample, it redistributes supervision according to the perceptual relevance of each region. This is achieved through a Perceptual Structuring Module (PSM) built on a pretrained vision backbone, which extracts perceptual relevance cues that describe the spatial and semantic structure of the generated content. These cues guide the advantage assignment during learning, without requiring dense annotations. In this way, ViPO performs fine-grained and 
spatial
selective credit assignment, allowing the model to focus updates on visually critical regions. 
This leads to more stable optimization, yielding improved perceptual fidelity and stronger alignment with human visual judgment across both image and video generation tasks.

The contributions of our work are summarized as follows:
\begin{itemize}[noitemsep, topsep=0pt, parsep=0pt, partopsep=0pt, leftmargin=1em]
\item We propose \textbf{Visual Preference Policy Optimization (ViPO)}, a redesigned GRPO framework for visual content generation. ViPO reformulates the advantage representation and assignment process, providing fine-grained and region-aware optimization suitable for both image and video generation.
\item We develop a \textbf{Perceptual Structuring Module (PSM)} that extracts perceptual relevance cues from pretrained vision backbones, enabling advantage redistribution without requiring pixel-level supervision or explicit region annotations.
\item We perform comprehensive experiments demonstrating that ViPO consistently surpasses vanilla GRPO, achieving stronger generalization, higher perceptual fidelity, and improved alignment with human visual judgment.
\end{itemize}
\section{Related Work}
\label{sec:related}
\noindent\textbf{RL for Visual Generation.}
Inspired by Proximal Policy Optimization (PPO)~\cite{ppo}, early works \cite{firstrlddpm,ddpo,fan2023dpok} integrated RL into diffusion models by optimizing the score function~\cite{scorebased} through policy gradient methods, thereby enabling the generation of images that better align with human preferences. 
Recently, GRPO-based approaches~\cite{dancegrpo,zhou2024flowgrpo,mixgrpo,tempflow} have pushed visual generation to new heights. In particular, DanceGRPO~\cite{dancegrpo} and FlowGRPO~\cite{zhou2024flowgrpo} adapt GRPO to visual generation by reformulating Flow Matching’s~\cite{flow_matching} ODE sampling into an SDE formulation, enabling online RL training on state-of-the-art visual generative models. To further improve efficiency, MixGRPO~\cite{mixgrpo} introduces a mixed ODE-SDE strategy with a sliding window mechanism, significantly reducing training overhead while maintaining performance. 
However, all these methods overlook the inherent characteristics of visual content, which, unlike language, possesses rich spatial dimensions that could be exploited for more fine-grained optimization.

\noindent\textbf{Visual Perception Modeling.} 
Modeling human visual perception has been a central theme in computer vision, with early approaches drawing direct inspiration from vision science. Saliency-based models~\cite{itti2001computational,jiang2015salicon} operationalized the idea that the visual system reduces scene complexity by prioritizing salient regions. Subsequent work~\cite{henderson2017meaning} highlighted the role of high-level semantics in guiding attention, leading to the notion of meaning maps, while eye-tracking studies~\cite{deepsaliency,henderson2018meaning} further revealed the non-uniform and dynamic nature of human gaze behavior. These perceptual insights have progressively shaped computational modeling, from the introduction of attention mechanisms in deep networks~\cite{visualattention}, to perceptual loss~\cite{perceptualloss} which explicitly measures discrepancies between CNN feature maps to approximate human perceptual similarity, and more recently to robotics~\cite{emulating}, where the adaptability of human vision inspired the Adaptive Vision Policy enabling agents to actively select optimal viewpoints.Visual preferences fundamentally rely on perceptual modeling. Building on this trajectory, we incorporate perceptual structuring into modern reinforcement learning for visual preference alignment, enabling content-adaptive optimization of visual content.

\noindent\textbf{Reward Model in Vision.}
A key bottleneck in applying RL to visual generation lies in the development of visual reward models. For image generation, recent works~\cite{pickscore,hpsv2,xu2023imagereward} have introduced preference-based reward models such as such as PickScore~\cite{pickscore}, HPSv2~\cite{hpsv2}, and ImageReward~\cite{xu2023imagereward}, which learn to predict human visual preferences. For video generation, VideoScore~\cite{he2024videoscore} introduces learnable metrics for direct evaluation, while VideoAlign~\cite{videoalign} assesses videos along three dimensions: visual quality, motion quality, and text alignment. More recently, VisionReward~\cite{visionreward} has been proposed as a fine-grained reward model for broader visual tasks. However, existing reward models primarily output scalar-level scores, which provide no information about where or why an image or video receives a high or low reward. More importantly, even though these models can capture fine-grained cues, a scalar reward collapses all spatial evidence into a single value. As a result, current GRPO-style alignment frameworks cannot exploit the rich spatial structure encoded in modern visual reward models.

To fully leverage these advances, we require a policy optimization mechanism that supports structured, interpretable, and spatially-aware optimization. Our goal is to develop such a framework, one that is compatible with a wide range of existing and future reward models.

\section{Method}
\label{sec:method}

We propose Visual Preference Policy Optimization (ViPO), an enhanced GRPO framework tailored for visual content generation. ViPO redefines both the \emph{advantage representation} and \emph{credit-assignment mechanism} of GRPO to better model the structured feedback inherent in images and videos. 
\begin{figure}
    \centering
    \setlength{\belowcaptionskip}{0pt}   
    \includegraphics[width=\linewidth]{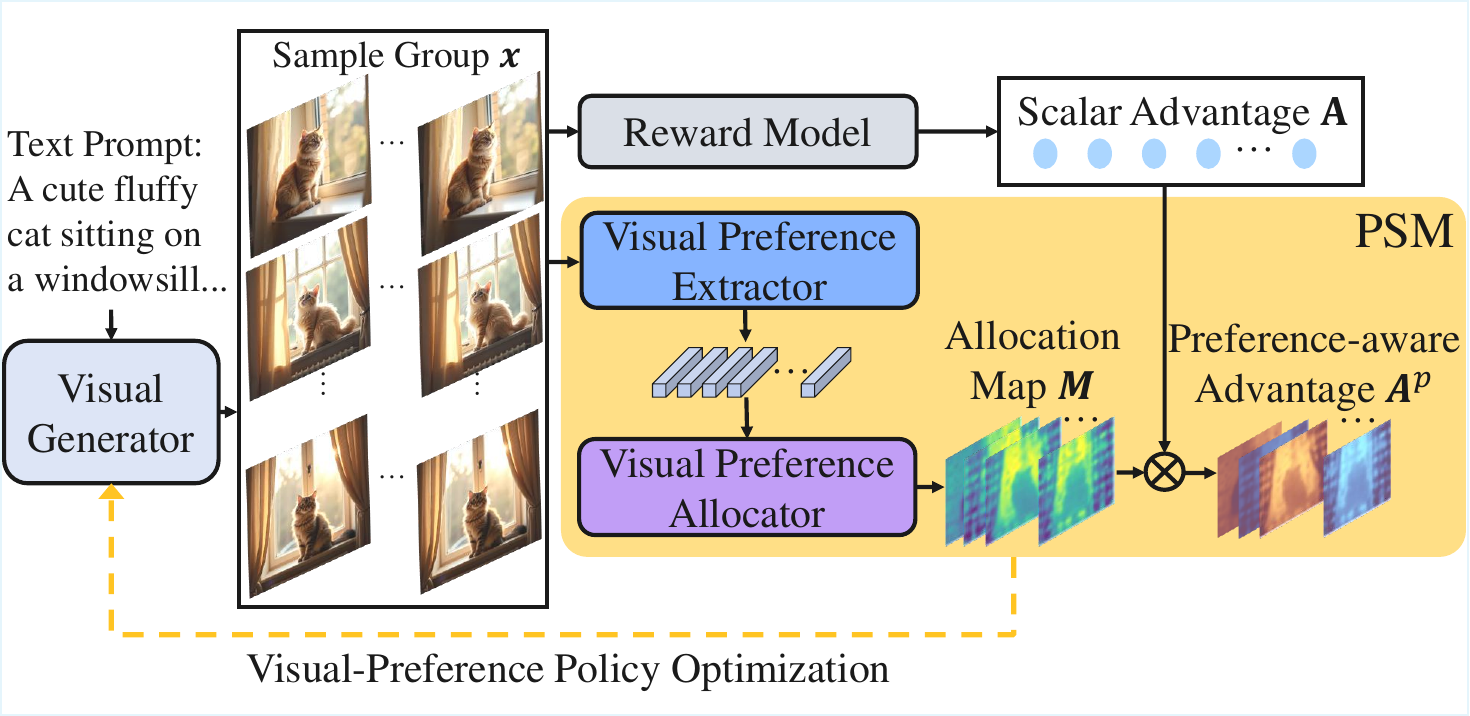}
    \caption{Overview framework of the proposed Visual Preference Policy Optimization (ViPO). Policy-sampled outputs are first evaluated by the reward model to obtain scalar advantages. In parallel, the samples are processed by the Perceptual Structuring Module (PSM) to produce allocation maps. The allocation maps are then combined with the scalar advantages to yield pixel-level, preference-aware advantages, which guide fine-grained visual preference policy optimization.}
    \label{fig:framework}
\end{figure}
While conventional GRPO computes a single scalar advantage per sample, ViPO introduces a Perceptual Structuring Module (PSM) that decomposes this global signal into region-aware weighting factors guided by visual preference cues. 
An overview of the ViPO framework is illustrated in Figure~\ref{fig:framework}. The standard group-wise reward computation of GRPO remains intact, but the resulting optimization pressure is redistributed across spatial and temporal dimensions according to perceptual relevance. This design allows ViPO to emphasize visually informative regions, yielding fine-grained alignment with perceptual preferences while maintaining the stability and simplicity of the original GRPO algorithm.

In this section, we first present the preliminaries of applying GRPO to visual generation, and then introduce our proposed Perceptual Structuring Module (PSM) and the full Visual Preference Policy Optimization.

\subsection{Preliminaries}
\label{sec:pre}
\noindent\textbf{GRPO for Visual Generation.}
The denoising process of the diffusion and rectified flow can be formulated as a Markov Decision Process (MDP). Thus GRPO~\cite{deepseek} can be applied as following. Given a prompt $c$, the generative policy will sample a group of outputs $\{o_1,o_2,...,o_G\}$ with a group size of $G$ and optimize the policy model by maximizing the following objective function:
\begin{equation}
\begin{split}
\mathcal{J}(\theta) &= \mathbb{E}_{\substack{\{\mathbf{o}_{i}\}_{i=1}^{G}\sim\pi_{\theta_{\mathrm{old}}}(\cdot|\mathbf{c}) \\ 
\mathbf{a}_{t,i}\sim\pi_{\theta_{\mathrm{old}}}(\cdot|\mathbf{s}_{t,i})}}
[\frac{1}{G}\sum_{i=1}^{G}\frac{1}{T}\sum_{t=1}^{T} \\
& \min(\rho_{t,i}A_{i}, \mathrm{clip}\left(\rho_{t,i},1-\epsilon,1+\epsilon)A_{i}\right)],
\end{split}
\end{equation}
where $\rho_{t,i}=\frac{\pi_{\theta}(\mathbf{a}_{t,i}\mid \mathbf{s}_{t,i})}
{\pi_{\theta_{\text{old}}}(\mathbf{a}_{t,i}\mid \mathbf{s}_{t,i})}$, $\pi_\theta(\mathbf{a}_{t,i}|\mathbf{s}_{t,i})$ is the policy function of MDP for output $\mathbf{o}_i$, and $A_i$ is the advantage function, computed using a group of rewards $\{r_1, r_2, ..., r_G\}$ correpsonding to the outputs within each group:
\begin{equation}
\label{eq:adv}
A_i=\frac{r_i-\textit{mean}(\{r_1,r_2,...,r_G\})}{\textit{std}(\{r_1,r_2,...,r_G\})}
\end{equation}

\noindent\textbf{SDE Sampling.}
State-of-the-art visual generative models increasingly adopt flow matching due to its efficiency and flexibility. However, flow matching typically relies on deterministic sampling based on an ordinary differential equation (ODE). The forward process in rectified flow~\cite{rectifiedflow} is defined as: $\mathrm{d}\mathbf{z}_t=\mathbf{u}_t\mathrm{d}t$, where $\mathbf{u}_t$ is the learned velocity field. The generative process reverses the ODE in time. However, GRPO requires stochastic exploration across multiple trajectory samples.
To support RL within flow-matching frameworks, it becomes necessary to convert the ODE formulation into a stochastic differential equation (SDE).

The corresponding reverse-time SDE can be written as:
\begin{equation}
\label{eq:sde}
\mathrm{d}\mathbf{z}_{t}=(\mathbf{u}_{t}-\frac{1}{2}\varepsilon_{t}^{2}\nabla\log p_{t}(\mathbf{z}_{t}))\mathrm{d}t+\varepsilon_{t}\mathrm{d}\mathbf{w},
\end{equation}
where $\varepsilon_{t}$ also introduces controlled stochasticity, and $\mathrm{d}\mathbf{w}$ denotes standard Brownian motion. Assuming the intermediate state $\mathbf{z}_{t}$ follows a Gaussian distribution $p_{t}(\mathbf{z}_{t})=\mathcal{N}(\mathbf{z}_{t}\mid\alpha_t \mathbf{x},\sigma_t^2I)$, the log-density term can be expressed as:
\begin{equation}
\log p_{t}(\mathbf{z}_{t})=\frac{-(\mathbf{z}_t-\alpha_t\mathbf{x})}{\sigma_t^2}
\end{equation}

Substituting this into the reverse SDE yields a tractable formulation for the conditional sampling policy $\pi_\theta(\mathbf{a}_{t}|\mathbf{s}_{t})$, enabling policy gradient optimization under the GRPO framework.

\subsection{Perceptual Structuring Module}
\label{sec:psm}
Human visual preference is inherently selective and spatially biased~\cite{desimone1995neural, itti2001computational, henderson2017meaning}: observers focus on semantically informative areas while discounting redundant background. To capture this property, ViPO introduces a Perceptual Structuring Module (PSM) that extracts visual preference cues and encodes them into a preference allocation map used for structured advantage assignment. The PSM comprises a Visual Preference Extractor (VPE) and a Visual Preference Allocator (VPA).

Given a generated image or video frame $\mathbf{x}\in\mathbb{R}^{H\times W\times 3}$, a visual preference extractor $\mathbf{\Phi}$ first produces feature embeddings that capture spatial organization and high-level semantics. The extractor outputs feature maps or patch embeddings denoted by $\mathbf{F}$. A dimensionality-reduction operator $\mathcal{R}(\cdot)$ (such as principal-component projection or eigen-space decomposition) is then applied to identify dominant feature directions and obtain a compact representation of visual preference:
\begin{equation}
\mathbf{Z} = \mathcal{R}(\mathbf{F}) \in \mathbb{R}^{N\times K},
\end{equation}
where $K$ denotes the number of retained components.  
The VPA then aggregates these components into a spatial map $\mathbf{S}\in\mathbb{R}^{H_p\times W_p}$ that reflects perceptual relevance. This fusion can be performed via variance-weighted summation:
\begin{equation}
\mathbf{S} = \operatorname{Reshape}\!\left(\sum_{j=1}^{K} \lambda_j z'_j\right),
\end{equation}
where $\lambda_j$ is the explained-variance ratio of the $j$-th component and $z'_j$ is its normalized projection. 
The map $\mathbf{S}$ is optionally smoothed and upsampled to the latent resolution, forming the final preference allocation map $\mathbf{M}$.  
For video, maps are computed per frame and temporally aligned to form a spatio-temporal volume $\mathbf{M}\in\mathbb{R}^{T_\ell\times H_\ell\times W_\ell}$.  
This process distills the structural relevance of each region without requiring dense labels or explicit annotations. The PSM thus serves as a 
bridge between perceptual feature distributions and policy optimization signals. Further implementation details on the choice of backbone extractors, the computation procedure, and the corresponding visualizations are provided in the supplementary material.

\subsection{Visual Preference Policy Optimization}
We now describe how ViPO incorporates the structured allocation map $\mathbf{M}$ into the policy optimization process.  
In standard GRPO, each generated sample $x_i$ receives a scalar advantage $A_i$. 
ViPO extends this formulation by distributing the advantage spatially and temporally. Let $p\in\mathcal{P}$ index a latent-space position across both spatial and temporal dimensions.  

The objective of ViPO is:
\begin{equation}
\begin{aligned}
\mathcal{J}(\theta)
&=\mathbb{E}\!\left[\frac{1}{G\,T_s\,|\mathcal{P}|}
\sum_{i=1}^{G}\sum_{t=1}^{T_s}\sum_{p\in\mathcal{P}}
\right.\\
&\qquad\left.
\min\!\Bigl(\rho_{t,i}^pA_i^p,\;
\operatorname{clip}(\rho_{t,i}^p,1-\epsilon,1+\epsilon)A_i^p\Bigr)\right],
\end{aligned}
\end{equation}
where $T_s$ denotes the number of diffusion or flow steps and $\rho_{t,i}^p$ is the local likelihood ratio.  
The spatially resolved advantage $A_i^p$ is defined as:
\begin{equation}
A_i^p = \mathbf{M}(p)\,A_i,
\end{equation}
linking the scalar group advantage $A_i$ with the regional weighting inferred by $\mathbf{M}$. Multiplying the allocation map with the advantage keeps the optimization direction consistent within each sample, prevents gradient interference from mixed-sign rewards, and preserves plug-and-play compatibility with existing GRPO implementations. This formulation provides fine-grained credit assignment and allows gradient updates to focus on perceptually significant regions across space and time.

In summary, ViPO enhances GRPO by introducing a PSM that extracts region-wise visual preference cues and by reformulating the policy objective to incorporate structured, region-weighted advantages.  
This approach maintains the theoretical simplicity and training stability of GRPO while improving its perceptual alignment and generative fidelity for both images and videos.
\section{Experiment}
\label{sec:experiment}
\subsection{Settings}
\noindent\textbf{Dataset.}
For image generation, we use the prompts from HPD~\cite{hpd}. The test set consists of 3200 prompts, encompassing four styles: “Animation”, “Concept Art”, “Painting”, and “Photo”. 
For video generation, we use the prompts from VidProM~\cite{vidprom} and randomly choose 1000 prompts as the test set, since VidProM does not provide a publicly released test split.

\noindent\textbf{Backbones and Rewards.} For image generation, we fine‑tune FLUX.1‑dev~\cite{blackforest2024flux} using HPSv2.1~\cite{hpsv2} as the reward model, and further assess out‑of‑domain (OOD) generalization with PickScore~\cite{pickscore} and ImageReward~\cite{xu2023imagereward}. For video generation, we fine‑tune Wan2.1‑T2V‑14B‑480P~\cite{wan2.1} with VideoAlign~\cite{videoalign}, which provides in‑domain reward signals for visual quality (VQ) and motion quality (MQ). OOD generalization is additionally evaluated on VBench~\cite{huang2024vbench}.

\noindent\textbf{Implement Details.}
For image generation, we use a group size of $G=12$ and downsample the training resolution to $512\times512$ with $8$ sampling steps.  
For video generation, we set the training resolution to $240\times416\times53$ ($H\times W\times T$), use a group size of $G=8$, and adopt $16$ sampling steps to accelerate training.  
During inference, the resolution and sampling steps are increased to $1024\times1024$ and $50$ for Flux and $480\times832\times53$ and $50$ for Wan2.1, respectively.  
All image generation experiments are conducted on $8\times$~NVIDIA H100 GPUs, while video generation experiments are trained on $32\times$~NVIDIA H100 GPUs.
Additional hyperparameter settings are provided in the supplementary material.
\begin{table}[t]
\caption{Quantitative comparison results of Flux. ViPO variants consistently outperform the original Flux model and DanceGRPO on both in-domain and out-of-domain metrics.}
\centering
\small
\setlength{\tabcolsep}{2.5pt}
\begin{tabular}{lccc}
\toprule
\multirow{2}{*}{{\centering \textbf{Method}}}
& \multicolumn{1}{c}{\textbf{In-domain}}
& \multicolumn{2}{c}{\textbf{Out-of-domain}} \\
\cmidrule(lr){2-2} \cmidrule(lr){3-4}
&\textbf{ HPSv2.1$\uparrow$} &\textbf{ PickScore$\uparrow$} & \textbf{ImageReward$\uparrow$} \\
\midrule
Flux      &0.3121 & 22.7038 & 1.1495  \\
DanceGRPO  &0.3203 & 22.5962 & 1.0392  \\
ViPO (DINO) & \textbf{0.3321} & \underline{22.8305} & \textbf{1.1883} \\
ViPO (SAM)     & 0.3219 & 22.6324 &1.1422   \\
ViPO (ResNet)     & \underline{0.3251} &\textbf{22.8492} &  \underline{1.1625} \\
\bottomrule
\end{tabular}
\label{tab:quant_flux}
\end{table}

\begin{table}[t]
\caption{Quantitative comparison results of Wan2.1. ViPO surpasses both the Wan2.1 and DanceGRPO in all out-of-domain criteria, demonstrating superior generalization.}
\centering
\small
\setlength{\tabcolsep}{2.5pt}
\begin{tabular}{lccccc}
\toprule
\multirow{2}{*}{\centering \textbf{Method}}
& \multicolumn{2}{c}{\textbf{In-domain}}
& \multicolumn{3}{c}{\textbf{Out-of-domain}} \\
\cmidrule(lr){2-3} \cmidrule(lr){4-6}
&\textbf{ VQ$\uparrow$} & \textbf{ MQ$\uparrow$} & \textbf{ Semantic$\uparrow$} & \textbf{Quality$\uparrow$} & \textbf{Total$\uparrow$}\\
\midrule
Wan2.1     &2.6219& 0.5896 &  83.36 & 71.20 & 80.92\\
DanceGRPO  & \underline{3.0935}&\underline{0.8639} & \underline{83.63}& 69.68 & 80.84  \\
ViPO &\textbf{3.5501} & \textbf{1.1515} & \textbf{83.98} & \textbf{72.59} & \textbf{81.70} \\
\bottomrule
\end{tabular}
\label{tab:quant_wan}
\end{table}





\begin{figure*}
    \centering
    \includegraphics[width=1\linewidth]{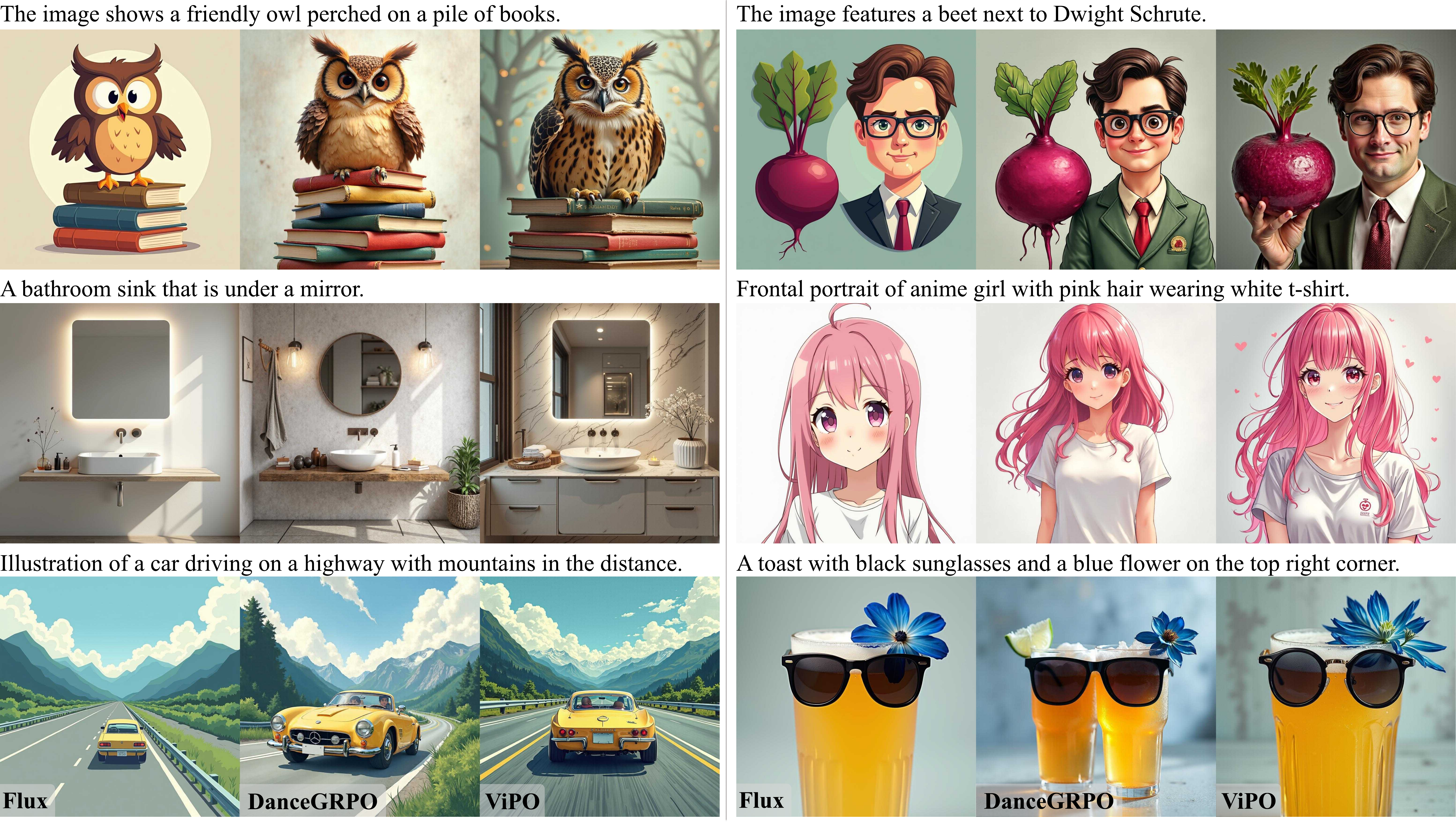}
    \caption{Qualitative comparison on Flux. Each group of results is arranged from left to right as follows: outputs from Flux, DanceGRPO, and our proposed ViPO. Our method demonstrates the best visual performance, exhibiting richer details, more realistic rendering, and overall superior perceptual quality.}
    \label{fig:flux_vis}
\end{figure*}

\begin{figure*}
    \centering
    \includegraphics[width=1\linewidth]{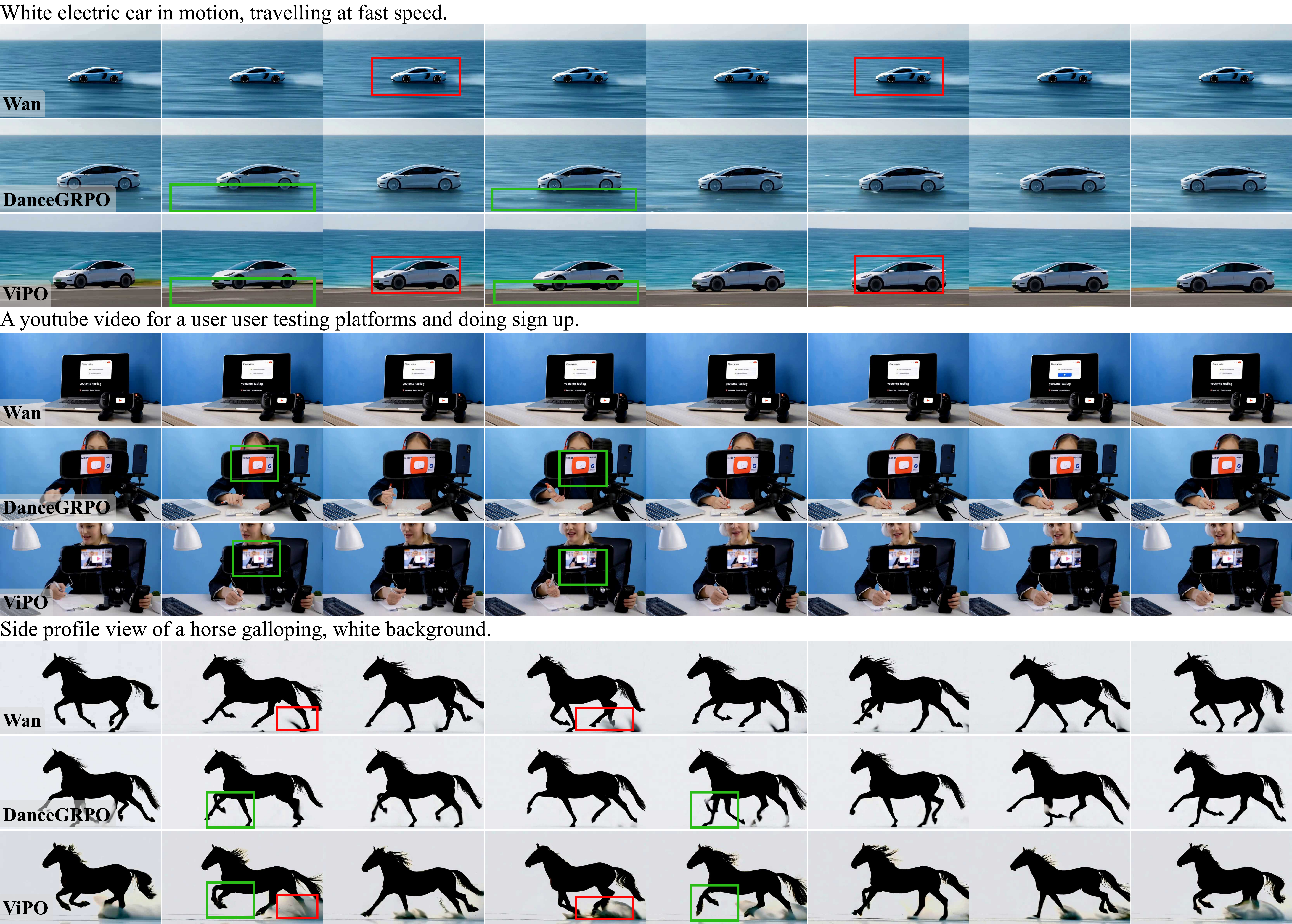}
    \caption{Qualitative comparison on Wan2.1. Each demo group is arranged top-to-bottom as follows: the result from Wan2.1, the output after applying DanceGRPO, and the output after applying ViPO. It is evident that our method delivers superior performance in terms of visual quality, and motion dynamics. In addition, we highlight representative regions with red boxes to indicate improvements over the Wan2.1, and green boxes to indicate improvements over DanceGRPO.}
    \label{fig:video_vis}
\end{figure*}

\begin{figure}
    \centering
    \includegraphics[width=1\linewidth]{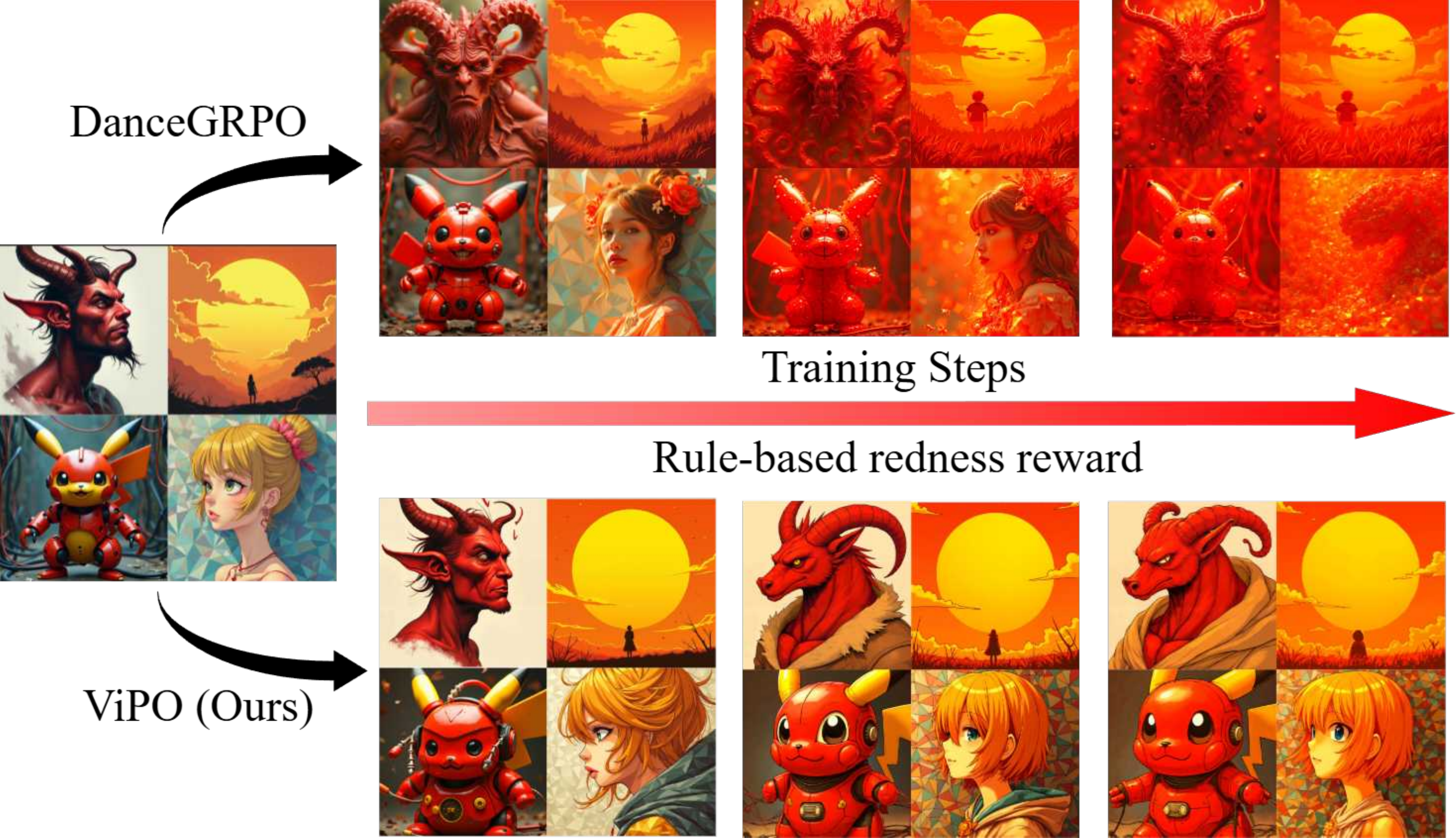}
    \caption{Comparison under the redness reward across training steps. As training progresses, results from DanceGRPO tend to suffer from semantic degradation and structural collapse. In contrast, ViPO consistently maintains the original semantic intent and structural integrity.}
    \label{fig:red}
\end{figure}

\subsection{Human Preference Reward}
\noindent\textbf{Quantitative Results.} 
To validate the effectiveness of the proposed Visual Preference Policy Optimization (ViPO) in both image and video generation, we conduct comprehensive quantitative and qualitative experiments under human preference–based reward models. 
As DanceGRPO~\cite{dancegrpo} represents one of the most recent and widely adopted GRPO-based methods for visual generation with diffusion and flow-matching models, we adopt it as the baseline to provide a rigorous and representative evaluation. In addition, we further examine the impact of different visual backbones within the PSM. 

The quantitative results of image generation are shown in Table~\ref{tab:quant_flux}. To assess the backbone sensitivity of ViPO, we construct three variants based on DINOv2~\cite{oquab2023dinov2}, SAM~\cite{sam}, and ResNet~\cite{resnet}, and all variants consistently outperform DanceGRPO across key metrics. Specifically, when HPS-v2.1 is used solely as the training reward model, ViPO achieves significant performance gains in both in-domain and out-domain evaluations. 
Among the variants, the DINO-based version performs the best, achieving the highest values in the in-domain HPSv2.1 and out-of-domain ImageReward. The ResNet-based variant exhibits unexpectedly good performance, particularly reaching the optimal value in the out-of-domain PickScore. The SAM-based variant is relatively weaker, but its metrics still surpass those of DanceGRPO. 
This may due to the features extracted by SAM being more inclined to low-level content rather than the high-level semantic information. 

For video generation, we exclusively adopt DINOv2 within the PSM to construct allocation maps. This choice is informed by our findings in the image generation, where DINOv2 consistently delivered the strongest semantic representations, and the variant built upon it achieved the best average performance. As reported in Table~\ref{tab:quant_wan}, ViPO surpasses both DanceGRPO and Wan2.1 in VQ and MQ, as well as out-of-domain VBench metrics including semantic, quality and overall scores.  Additional details of the VBench results across different dimensions are provided in the supplementary material. Since DanceGRPO did not initially provide an official implementation for Wan2.1, we use our own implementation for this comparison.

Across both image and video generation, ViPO consistently improves in in-domain metrics and achieves gains under out-of-domain evaluation. This shows that structured, region-aware preference cues provide a more informative optimization signal than conventional scalar feedback. By redistributing the learning pressure according to perceptual relevance, ViPO enhances both fidelity and robustness under distribution shifts, confirming the effectiveness of perceptual structuring for preference-aligned visual generation.



\noindent\textbf{Qualitative Results.} 
Figure~\ref{fig:flux_vis} presents qualitative comparisons among the original Flux, DanceGRPO, and ViPO. ViPO consistently produces more detailed, realistic, and preference-aligned results. For instance, in the first row's rightmost example, although DanceGRPO introduces more visual detail, the beet appears unnaturally placed beside the man. By comparison, ViPO not only renders both the man and the beet more realistically, but also depicts the man holding the beet, which aligns better with real-world semantics. Similarly, in the third row's rightmost example, DanceGRPO adds background detail but duplicates the foreground glass. ViPO enhances background while preserving foreground coherence.

Figure~\ref{fig:video_vis} presents qualitative results for video generation. As shown, our method significantly improves both visual fidelity and motion quality, consistent with the quantitative gains observed in VBench metrics. In the top example, both DanceGRPO and ViPO enhance camera perspective, but ViPO further refines the rendering of the white electric car and the road surface, yielding results that better align with human aesthetic and physical plausibility. In the middle example, GRPO-based optimization generally produces more detailed and complex frames; however, compared to DanceGRPO, ViPO generates more realistic screen content, as it captures the background person in a way similar to smartphone photography, thereby enriching scene authenticity. In the bottom example, ViPO demonstrates clear advantages in dynamic realism: the running horse exhibits stronger and more natural motion, with fluid water splashes and no structural artifacts. By contrast, DanceGRPO increases motion amplitude but introduces semantic distortions such as duplicated or partially broken legs. 

These qualitative improvements can be attributed to the proposed PSM. By decomposing perceptual features into spatially organized preference maps, the PSM enables reward attribution to be concentrated on regions that are more aligned with human visual preference. This region-differentiated optimization allows ViPO to apply varying degrees of refinement across different areas, focusing on semantically meaningful structures such as dynamic motion or fine-grained details, rather than performing uniform updates over the entire frame. In contrast, GRPO’s scalar-wise global optimization can propagate misleading gradient signals to inappropriate regions, which sometimes results in subtle structural artifacts—for example, duplicated or broken limbs in the running horse. By selectively allocating optimization strength, ViPO mitigates such issues and produces outputs that are both visually coherent and semantically aligned. More examples and visual comparisons can be found in the supplementary material.



\subsection{Redness Reward}
We also conduct experiments using a rule-based reward function. Specifically, we adopt a redness reward function $r(x)$, which is defined as the difference between the red channel intensity and the average of the green and blue channel intensities: 
\begin{equation}
r(x) = x^0 - \frac{1}{2}(x^1+x^2),    
\end{equation}
where $x^i$ denotes the $i$-th channel of the generated frame. 

The results are illustrated in Figure~\ref{fig:red}. As training progresses, DanceGRPO tends to degrade the semantic content of the generated outputs. For example, in the bottom-right case, the girl eventually collapses into an unrecognizable shape in the final training step. In comparison, our method preserves the semantic integrity throughout training. Even in the bottom-right example, where the girl's hair and background turn red due to the reward signal, the overall structure and identity remain intact. This also indicates that our visual preference-guided, region‑differentiated optimization is less susceptible to collapse under global gradient signals, thereby better preserving semantic integrity even when color channels are strongly biased.

\subsection{Ablation Study}
To better understand the design and influential factors of the proposed Perceptual Structuring Module (PSM), we conduct a series of ablation studies on the Flux model, as summarized in Table~\ref{tab:ablation_aggregation} and Table~\ref{tab:ablation_number}. Our analysis focuses on four key components of the PSM: (1) the necessity of the visual preference allocation map, (2) the aggregation strategy used in the Visual Preference Allocator (VPA), (3) the number of principal components retained in the Visual Preference Extractor (VPE), and (4) the effect of spatial smoothing applied in the VPA. These studies provide insight into how each design choice contributes to the effectiveness and stability of ViPO.

\noindent\textbf{Visual Preference Allocation Map.}
Replacing the visual preference allocation map with an all-ones map leads to a clear performance drop. Although this setting is theoretically equivalent to original GRPO, the pixel-wise formulation introduces additional variance when the allocation map lacks semantic structure. This confirms that the benefit of our method comes from semantically meaningful fine-grained allocation guided by perception mechanism rather than pixel-wise decomposition alone. 

Moreover, applying the allocation map directly to the reward instead of the advantage also degrades performance. Because semantic regions vary across samples, so the same concept may appear at different locations with different weights, producing mismatched advantages. Within a single sample, it can assign conflicting gradients to the same object, disrupting optimization. By contrast, applying the map on the advantage preserves stable relative signals while still enabling fine-grained semantic allocation.

\noindent\textbf{Aggregation Ways.}
To aggregate the principal components derived from VPE, we evaluate two schemes: simple averaging and variance-weighted aggregation. The averaging baseline treats all components equally, implicitly assuming equal semantic contribution across components. The variance-weighted approach, instead, assigns higher weights to components that explain more variance, thereby emphasizing directions that capture stronger semantic signals. Empirically,  the variance-weighting yields higher out-of-domain scores across benchmarks. This indicates that prioritizing components with greater explanatory power provides a more faithful representation of semantic importance, while uniform averaging may dilute the contribution of informative components by mixing them with less relevant directions. These results highlight the role of aggregation in bridging low-level feature decomposition with high-level preference alignment.

\noindent\textbf{Number of Principal Components.}
We vary the number of retained PCA components $K$ from 1 to 5 and observe modest, metric-dependent gains rather than a strictly monotonic trend. HPS score improves up to $K=4$, ImageReward peaks at $K=2$, and PickScore slightly favors $K=5$, indicating that adding components beyond $K=3$ starts to capture weaker directions that help one metric while marginally hurting others. Across metrics, $K=3$ offers a robust balance, competitive HPS, strong ImageReward and stable PickScore, without the variability seen when more components are included. In addition, retaining three components provides good interpretability, since they can be projected into the RGB color space for visualization. We therefore adopt $K=3$ as the default, prioritizing semantic coverage and stability over marginal, metric-specific gains.

\noindent\textbf{Effect of Spatial Smoothing.}
We also study the Gaussian smoothing strength $\sigma$ applied to the allocation map. From the Table~\ref{tab:ablation_number}, we find that removing smoothing still yields competitive results, indicating that the allocator remains effective even without this step. However, smoothing generally improves robustness across metrics, while overly aggressive kernels ($\sigma=2$) degrade performance. A moderate kernel ($\sigma=1$) provides the most consistent balance, and we adopt it as the default while noting that the unsmoothed variant remains a viable alternative. Intuitively, since the feature maps extracted by the VPE may contain local jitter or noisy activations when projected into spatial maps, applying Gaussian smoothing helps regularize these fluctuations and yields more stable preference allocation.
\section{Conclusion}
\label{conclusion}

\begin{table}[t]
\caption{Ablation study on allocation map and aggregation strategies.}
\setlength{\tabcolsep}{1.5pt}
\centering \small
\begin{tabular}{lccc}
\toprule
\textbf{Method} & \textbf{HPSv2.1$\uparrow$} & \textbf{PickScore$\uparrow$} & \textbf{ImageReward$\uparrow$} \\
\midrule
\multicolumn{4}{@{}l}{\textit{Allocation Map}} \\
Uniform (all ones) & 0.3043 & 22.2043 & 0.9520 \\
Reward map & 0.3090 & 22.3866 & 1.0058 \\
Advantage map & \textbf{0.3321} & \textbf{22.8305} & \textbf{1.1883} \\
\midrule
\multicolumn{4}{@{}l}{\textit{Aggregation Strategy}} \\
Average & 0.3238 & 22.7037 & 1.1318 \\
Weighted & \textbf{0.3321} & \textbf{22.8305} & \textbf{1.1883} \\
\bottomrule
\end{tabular}
\label{tab:ablation_aggregation}
\end{table}
\begin{table}[t]
\caption{Ablation study on number of principal components and spatial smoothing.}
\setlength{\tabcolsep}{2pt}
\centering
\small
\begin{tabular}{l|cccc}
\toprule
\multicolumn{2}{c}{\textbf{HyperParams}} 
    & \textbf{HPSv2.1$\uparrow$} 
    & \textbf{PickScore$\uparrow$} 
    & \textbf{ImageReward$\uparrow$} \\
\midrule

\multirow{5}{*}{$K$}
 & $1$ & 0.3291 & 22.9286 & 1.1537 \\
 & $2$ & 0.3260 & 22.7441 & 1.2155 \\
 & $3$ & 0.3321 & 22.8305 & 1.1883 \\
 & $4$ & 0.3337 & 22.8025 & 1.1862 \\
 & $5$ & 0.3273 & 22.9324 & 1.1925 \\
\midrule

\multirow{5}{*}{$\sigma$}
 & w/o smooth & 0.3325 & 22.7996 & 1.1618 \\
 & $0.5$ & 0.3059 & 22.4758 & 0.9527 \\
 & $1$   & 0.3321 & 22.8305 & 1.1883 \\
 & $1.5$ & 0.3305 & 22.5958 & 1.1828 \\
 & $2$   & 0.3204 & 22.7703 & 1.1466 \\
\bottomrule
\end{tabular}
\label{tab:ablation_number}
\end{table}

In this paper, we introduced Visual Preference Policy Optimization (ViPO), a pixel-wise RL framework inspired by human visual preferences that integrates perceptual structuring into GRPO. By redistributing optimization pressure toward perceptually important regions, ViPO enhances semantic integrity and achieves stronger alignment with human preference. Besides, ViPO provides a modular and lightweight framework bridging perceptual modeling with RL, fully compatible with existing GRPO pipelines. Looking ahead, its spatial awareness and differentiated assignment suggest promising directions for future research, including structured feedback, region‑aware policy learning, and perceptual alignment in high‑dimensional generative tasks.
{
    \small
    \bibliographystyle{ieeenat_fullname}
    \bibliography{main}
}
\clearpage
\setcounter{page}{1}
\maketitlesupplementary
\renewcommand{\thesection}{\Alph{section}}
\setcounter{section}{0}

%
Due to the page constraint of the main paper, additional methodological details, parameter settings, and extended qualitative and quantitative results are provided in the supplementary material. The content is organized into the following parts:
\begin{itemize}
    \item The computation of features corresponding to different backbone choices within Perceptual Structruring Module (PSM), along with visualizations of the allocation maps and the results from different ViPO variants.
    \item More quantitative and qualitative comparisons.
    \item The hyperparameter configurations used during training.
\end{itemize}

\begin{figure*}
    \centering
    \includegraphics[width=1\linewidth]{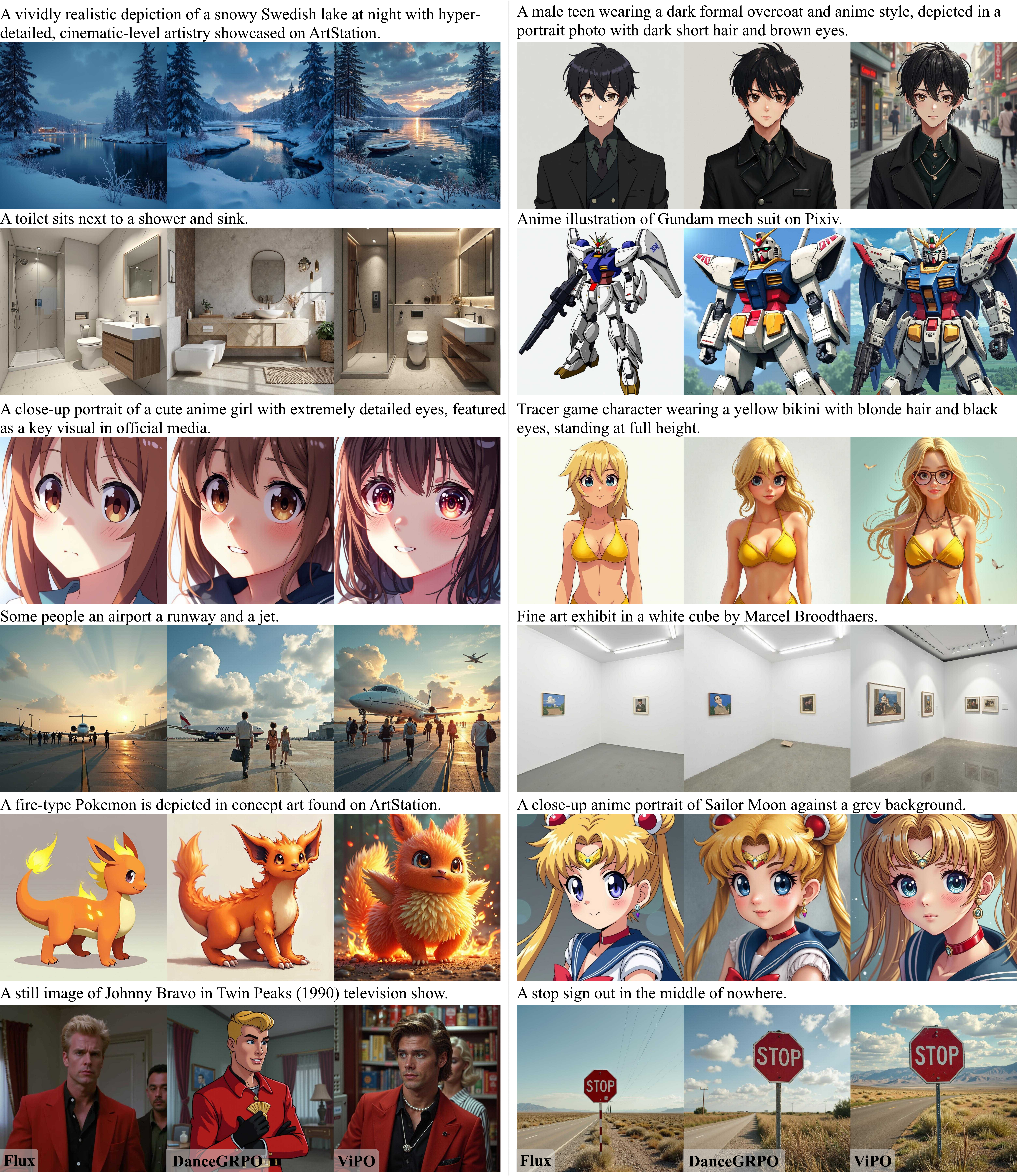}
    \caption{More qualitative comparison results for Flux. Each group of images, from left to right, shows the output from Flux, DanceGRPO, and our ViPO. As observed, DanceGRPO optimization generally introduces richer details and improved composition. In contrast, ViPO achieves more refined enhancements, delivering superior visual quality and finer improvements.}
    \label{fig:sup_flux}
\end{figure*}
\begin{figure*}
    \centering
    \includegraphics[width=1\linewidth]{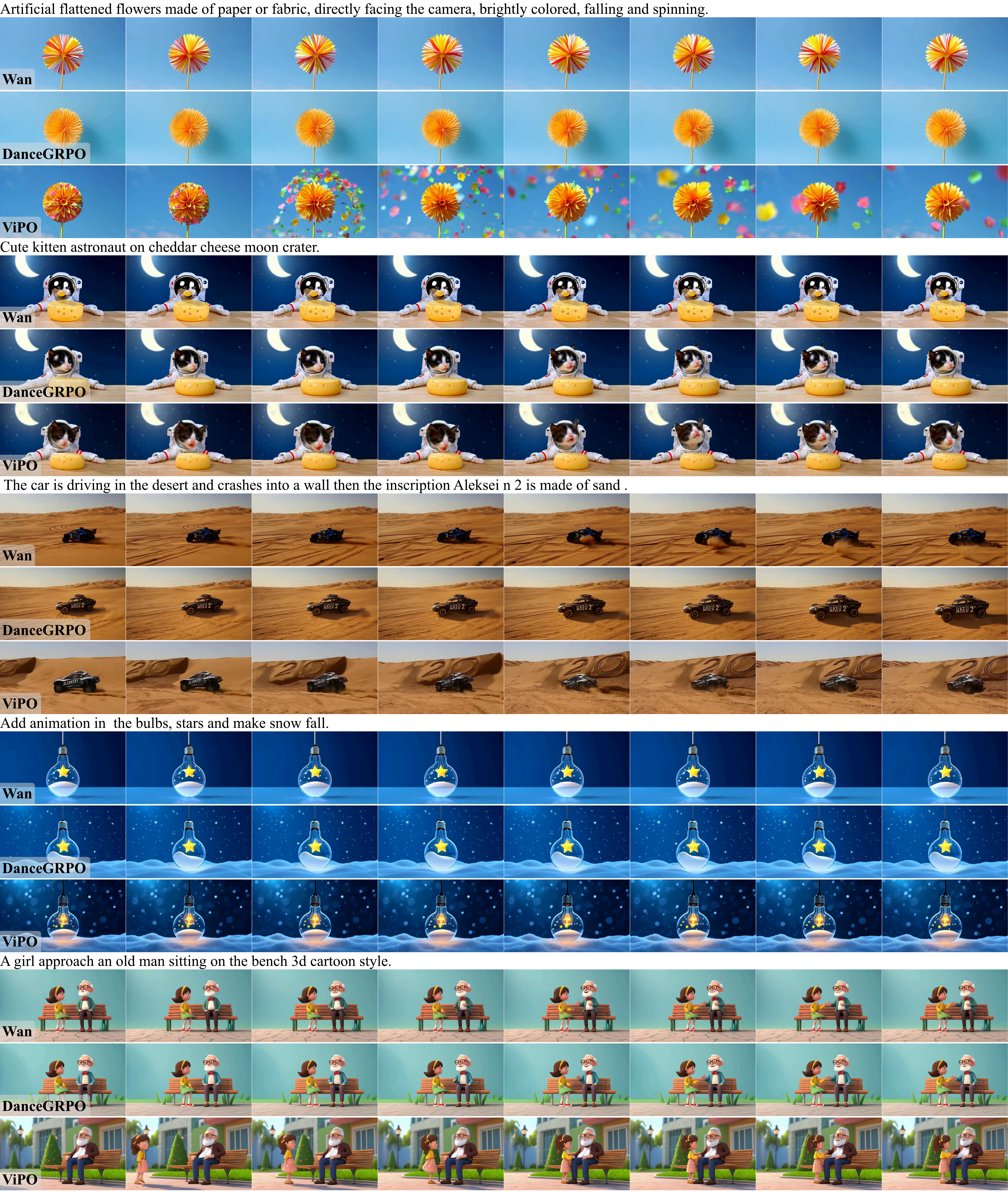}
    \caption{More qualitative comparison of video generation. For each group of sequences, the rows correspond to outputs from Wan2.1, DanceGRPO, and ViPO, respectively. As shown, DanceGRPO tends to enhance visual detail and yields moderate improvements in dynamic fidelity. In contrast, ViPO achieves more substantial gains in motion quality and visual realism, while further strengthening semantic alignment with the prompts.}

    \label{fig:sup_wan}
\end{figure*}

\section{Details of PSM}
\textbf{Computation Details.}
The Perceptual Structuring Module (PSM) enriches the scalar reward signal by modeling human visual preference through pretrained vision backbones. As detailed in Section~\ref{sec:psm}, PSM extracts perceptual features and constructs allocation maps that guide the redistribution of advantages across spatio-temporal locations. In the experiments, DINOv2~\cite{oquab2023dinov2}, ResNet~\cite{resnet}, and SAM~\cite{sam} are respectively adopted as perceptual backbones for evaluating the effectiveness of PSM.

For Transformer-based perception models such as DINOv2 and SAM, 
$\mathbf{\Phi}$ outputs patch-level features $\mathbf{F} \in \mathbb{R}^{N \times D}$, 
where $N = H_p \times W_p$, $H_p = H/p$, $W_p = W/p$, $p$ is the patch size, and $D$ is the feature dimension. 
Because each patch embedding is semantically homogeneous, Principal Component Analysis (PCA) is applied to
$\mathbf{F}$ to retain the top-$K$ components:
\begin{equation}
	\mathbf{Z} = \operatorname{PCA}(\mathbf{F}) \in \mathbb{R}^{N \times K}, \quad
	\boldsymbol{\lambda} = (\lambda_1, \lambda_2, \dots, \lambda_K),
\end{equation}
where $\lambda_j$ denotes the explained variance ratio of the $j$-th component. Each column $z_j \in \mathbb{R}^N$ represents the $j$-th principal component. 

PCA decomposes the embedding space into semantic directions, providing a compact representation of visual importance. For each component $z_j$, values are normalized and inverted so that regions with lower PCA projections (often corresponding to salient content) receive higher importance:
\begin{equation}
z'_j = \frac{\max(z_j) - z_j}{\max(z_j) - \min(z_j)}, \quad j = 1,2,\dots,K.
\end{equation}

The aggregated semantic map $\mathbf{S}$ is obtained by reshaping the PCA-projected features into a 2D map. A variance-weighted scheme is commonly used:
\begin{equation}
\mathbf{S} = \operatorname{Reshape}\left(\sum_{j=1}^{K} \lambda_j z'_j\right) \in \mathbb{R}^{H_p \times W_p}
\end{equation}

For CNN-based models~\cite{resnet}, intermediate feature maps $\mathbf{F} \in \mathbb{R}^{C \times H_f \times W_f}$ are extracted from a designated layer, where $\mathbf{F}_c \in \mathbb{R}^{H_f \times W_f}$ denotes the $c$-th channel. An activation map $\mathbf{S} \in \mathbb{R}^{H_f \times W_f}$ is obtained via channel-weighted aggregation:
\begin{equation}
\mathbf{S} = \sum_{c=1}^{C} \alpha_c \mathbf{F}_c,
\label{eq:activation_map}
\end{equation}
where the weights $\boldsymbol{\alpha} = [\alpha_1, \ldots, \alpha_C] \in \mathbb{R}^C$ are derived from global average pooling followed by softmax:
\begin{equation}
\boldsymbol{\alpha} = \operatorname{softmax}\!\left(\frac{1}{H_f W_f} \sum_{i=1}^{H_f} \sum_{j=1}^{W_f} \mathbf{F}_{:,i,j}\right)
\label{eq:channel_weight}
\end{equation}

The resulting map $\mathbf{S}$ is further optionally smoothed and upsampled to the latent spatial resolution, yielding the final allocation map $\mathbf{M}$ used in PSM.

In our experiments, ResNet-50 are adopted as the CNN backbone and extract features from \texttt{layer4}, 
since this layer provides a good balance between semantic richness and spatial resolution, making it suitable for constructing allocation maps.

\noindent\textbf{Training Overhead.}
To further quantify the computational overhead introduced by PSM, we report the training overhead of ViPO with different perceptual backbones in Table~\ref{tab:overhead}. PSM introduces only minor overhead relative to the original GRPO training pipeline: the training step time increases by only 1.0--1.8\% for image generation and 4.8\% for video generation, while the increase in peak memory usage remains below 4.5\%. The detailed cost analysis (forward / PCA / resample) indicates that PSM incurs negligible latency relative to the total training step, without modifying the generative backbone.
\begin{table}[t]
	\centering
	\caption{Training overhead of ViPO with different PSM backbones, compared with the corresponding DanceGRPO baseline.}
	\label{tab:overhead}
	\scriptsize
	\setlength{\tabcolsep}{4pt}
	\resizebox{0.48\textwidth}{!}{
		\begin{tabular}{lccc}
			\toprule
			\textbf{Method} & \textbf{Step Time (s)} & \textbf{PSM/sample (ms)} & \textbf{Peak Mem (MB)} \\
			\midrule
			Flux+DanceGRPO       & 59.66 & -- & 47084 \\
			Flux+ViPO (DINO)     & 60.47\textcolor{gray}{ (+1.4\%)} & 49.73\textcolor{gray}{ (27.9/21.4/0.5)} & 48964\textcolor{gray}{ (+4.0\%)} \\
			Flux+ViPO (ResNet)   & 60.24\textcolor{gray}{ (+1.0\%)} & 4.22\textcolor{gray}{ (3.7/0.2/0.4)} & 47200\textcolor{gray}{ (+0.3\%)} \\
			Flux+ViPO (SAM)      & 60.75\textcolor{gray}{ (+1.8\%)} & 33.25\textcolor{gray}{ (20.5/12.2/0.6)} & 47422\textcolor{gray}{ (+0.7\%)} \\
			\midrule
			Wan+DanceGRPO        & 203.45 & -- & 25498 \\
			Wan+ViPO (DINO)      & 213.27\textcolor{gray}{ (+4.8\%)} & 972.5\textcolor{gray}{ (413/498/17.5)} & 26654\textcolor{gray}{ (+4.5\%)} \\
			\bottomrule
		\end{tabular}
	}
\end{table}

\begin{figure}
    \centering
    \includegraphics[width=1\linewidth]{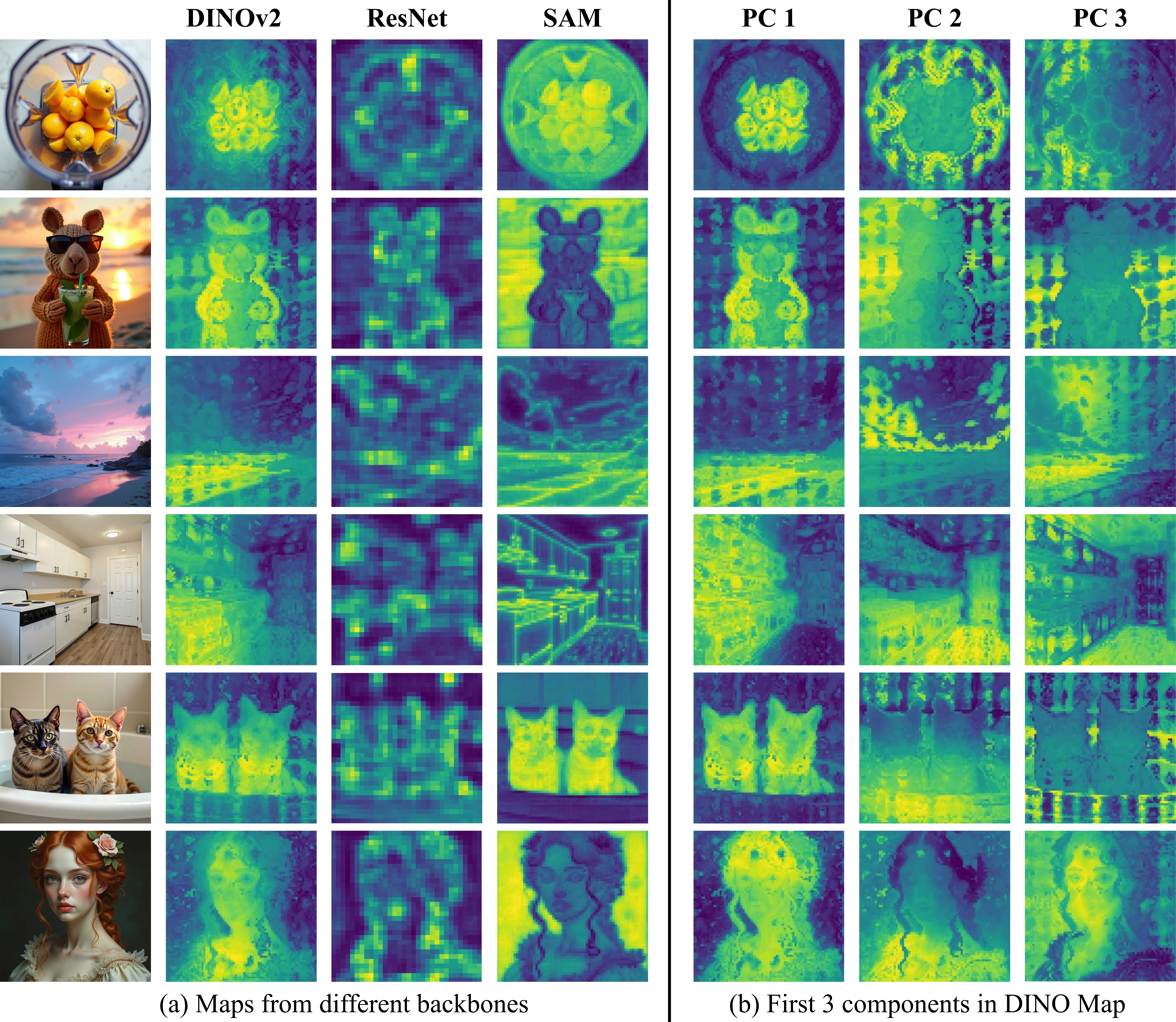}
    \caption{Visualization of allocation maps. (a) Allocation maps produced by the PSM with different vision backbones. From left to right: the original image generated by Flux, followed by maps obtained using DINOv2, ResNet, and SAM. (b) Visualization of the top three principal components that compose the allocation map derived from DINOv2.}
    \label{fig:map}
\end{figure}

\noindent\textbf{Visualization Examples.} Additional visualizations are provided to illustrate the behavior of PSM under different perception backbones. Figure~\ref{fig:map} (a) shows the final allocation maps generated by DINOv2, ResNet, and SAM, respectively. Figure~\ref{fig:map} (b) further visualizes the first three principal components extracted from DINOv2 embeddings. In both subfigures, brighter colors indicate higher allocation weights.

It can be observed that DINOv2 produces maps that closely align with human visual preference, often assigning higher weights to semantically salient regions that typically attract immediate human attention, such as primary objects or dominant visual entities.
In contrast, ResNet tends to respond to texture and spatial structure, yielding more distributed activations. SAM emphasizes boundary-aware regions, focusing on contours and segmentable areas as shown in the third and forth rows of Figure~\ref{fig:map} (a).
Moreover, for maps derived by SAM, high allocation weights do not always correspond to semantic foreground. For example, in the second and sixth rows, background regions receive stronger emphasis, while in the fifth row, the cat is clearly highlighted.

Figure~\ref{fig:map} (b) visualizes the first three principal directions extracted from DINOv2 embeddings. Each component corresponds to a distinct semantic region, and their ordering generally reflects the progression of visual saliency, beginning with primary objects and extending to secondary structures or contextual cues. These components are subsequently integrated through a weighted aggregation to construct the final allocation map, enabling PSM to highlight perceptually meaningful regions in a manner aligned with human visual preference.

\begin{figure}
    \centering
    \includegraphics[width=1\linewidth]{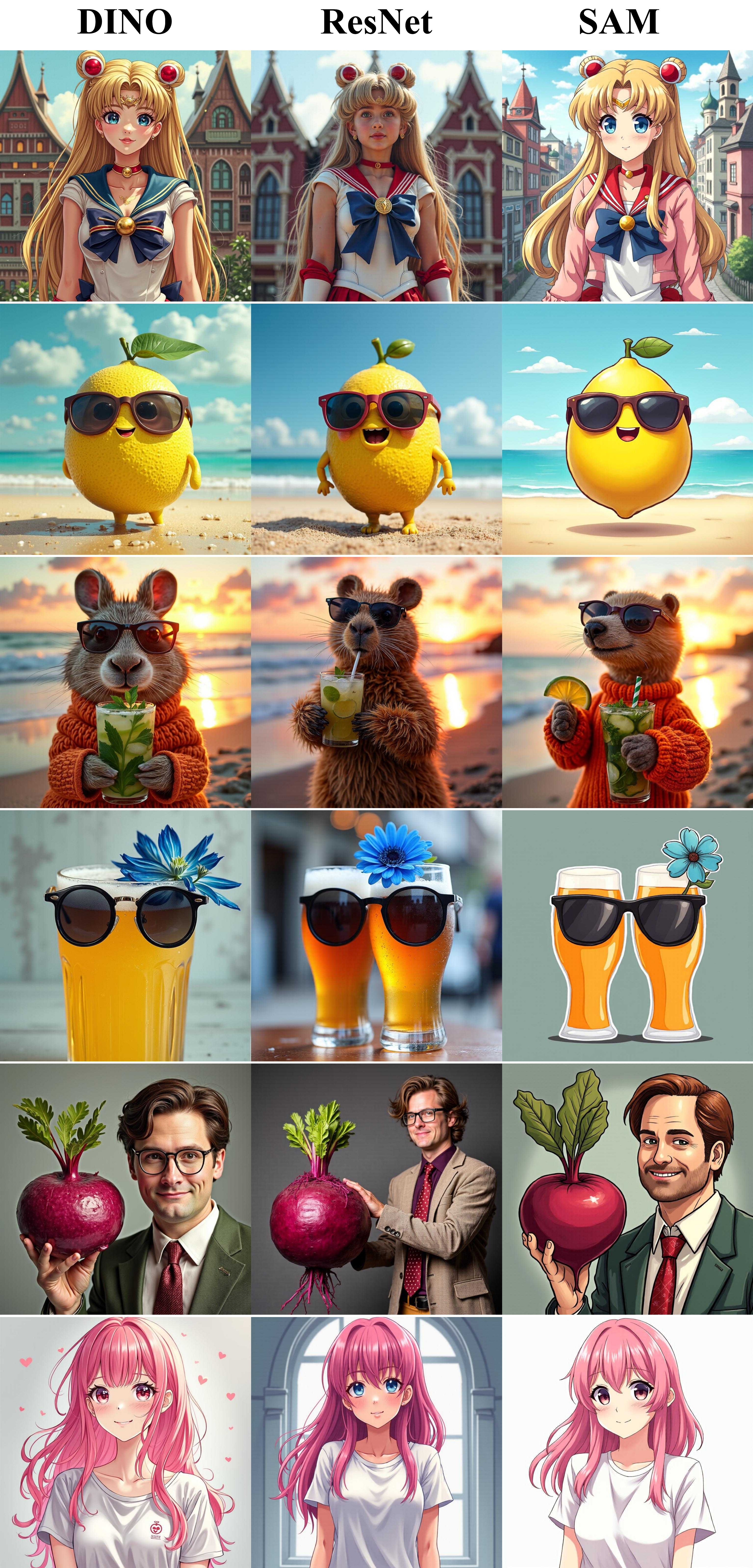}
    \caption{Visualization of results obtained with different ViPO variants. From left to right: ViPO with DINOv2 as the PSM backbone, ViPO with ResNet, and ViPO with SAM.}
    \label{fig:abla}
\end{figure}

Figure~\ref{fig:abla} presents results from ViPO variants trained on Flux with different perception backbones. The comparisons show that backbone choice influences how advantages are adaptively allocated according to visual content, which in turn affects generation quality and further supports the effectiveness of perceptual structuring in preference-aware optimization.

\section{More Results}
\begin{table}[t]
\caption{Quantitative comparison across detailed evaluation dimensions in VBench. ViPO consistently achieves superior performance across most dimensions.}
\centering
\small
\setlength{\tabcolsep}{4pt}
\begin{tabular}{lccc}
\toprule
\textbf{VBench} & \textbf{Wan2.1} & \textbf{DanceGRPO} & \textbf{ViPO} \\
\midrule
Dynamic Degree & 52.77 & 45.83 & \textbf{63.89} \\
Imaging Quality & 67.90 & \underline{68.31} & \textbf{68.88} \\
Multiple Objects & 69.96 & 63.18 & \textbf{74.70} \\
Color & 89.20 & 86.51 & \underline{88.97} \\
Spatial Relationship & 72.94 & 71.18 & \textbf{81.44} \\
Temporal Style & 24.12 & \underline{24.14} & \textbf{24.25} \\
Appearance Style & 21.51 & 20.91 & \underline{21.39} \\
Scene & 32.48 & 29.14 & \underline{31.90} \\
Overall Consistency & 25.06 & 25.02 & \textbf{25.32} \\
\bottomrule
\end{tabular}
\label{tab:quant_vb16_transposed}
\end{table}
In this section, more detailed evaluation results on Wan2.1 are provided. As shown in Table~\ref{tab:quant_vb16_transposed}, ViPO achieves a substantial improvement on \textit{Dynamic Degree}, and also yields gains in \textit{Imaging Quality}, which is consistent with the qualitative results. Furthermore, several semantics-related dimensions, including \textit{Multiple Objects}, \textit{Spatial Relationship}, and \textit{Temporal Style}, are improved. The enhancement in \textit{Overall Consistency} further demonstrates the superiority of our approach.

Additional qualitative visualizations are also provided. Figure~\ref{fig:sup_flux} illustrates extended results on Flux, where ViPO consistently produces outputs with richer details and improved aesthetics. Figure~\ref{fig:sup_wan} shows extended qualitative comparisons on video generation, where ViPO achieves noticeable improvements in motion fidelity, visual quality, and semantic alignment. Interestingly, Although the Text Alignment score is not explicitly included as a reward signal, semantic alignment still shows improvement. likely as a side effect of optimizing for different regions of visual preference, which indirectly enhances semantic consistency. More qualitative comparisons for both video and image results are included in the supplementary MP4 file.

\begin{figure}
    \centering
    \setlength{\belowcaptionskip}{0pt}
    \includegraphics[width=1\linewidth]{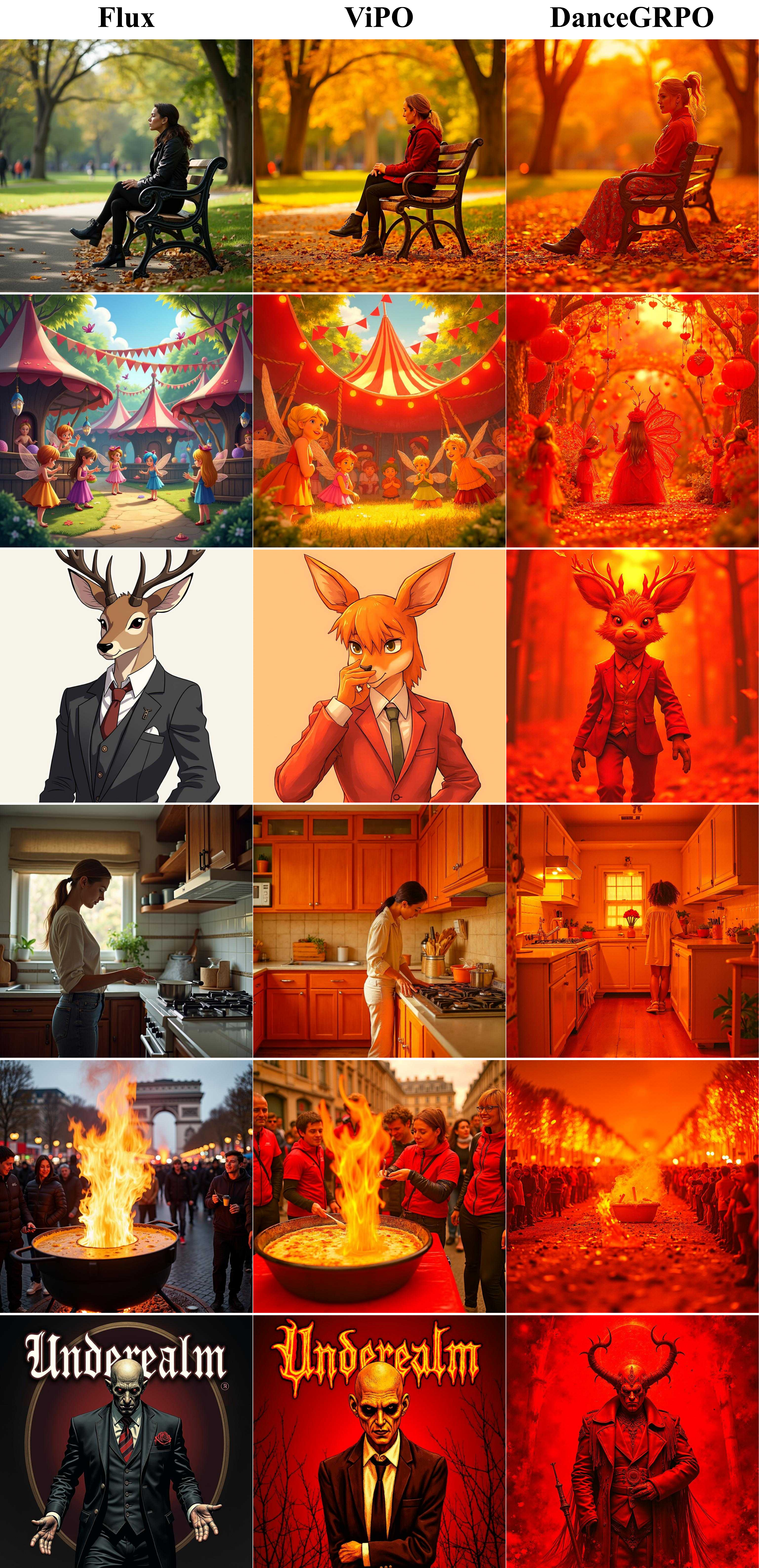}
    \caption{More comparison of results using the redness reward. The left column shows outputs from Flux without RL fine-tuning, the middle column presents results from ViPO, and the right column displays results from DanceGRPO. The comparisons indicate that ViPO better preserves the semantic content of the images, maintaining object identity and visual coherence under preference optimization.}
    \label{fig:red_more}
\end{figure}

In addition, further qualitative visualizations based on the redness reward are provided in Figure~\ref{fig:red_more}. The results show that ViPO better preserves the semantic content of the images compared to baselines, maintaining object identity and visual coherence under preference optimization. These results further confirm the effectiveness of our approach in aligning visual generation with human visual preference.

\section{Training Details}
The parameter $\eta$ controls the level of randomness in SDE sampling. In the reverse-time SDE formulation (Equation~\ref{eq:sde}), the stochastic term is instantiated as $\varepsilon_t = \eta \sqrt{\Delta t}$, where $\Delta t$ denotes the step size in the noise schedule. For Flux, $\eta$ is set to $0.3$, while for Wan2.1 it is set to $0.25$. The learning rate was configured as $1 \times 10^{-5}$ for Flux and $5 \times 10^{-6}$ for Wan2.1. During training, backpropagation is not performed through all sampling steps; instead, a timestep fraction of $0.6$ is used, meaning that only $60\%$ of the timesteps contribute to gradient updates. All samples within a group are generated from the same initialization noise to ensure consistency. In the training objective, the clipping range for the importance ratio $\rho^p$ is set to $1 \times 10^{-4}$. For Wan2.1, videos are sampled with 53 frames at 16 FPS, while the reward model processes them at 2 FPS.



\end{document}